\pdfoutput=1

\documentclass[11pt]{article}

\usepackage{ACL2023}

\usepackage{times}
\usepackage{latexsym}
\usepackage{graphicx}
\usepackage{amsmath}
\usepackage{amssymb}
\usepackage{bbm}
\usepackage{bm}

\usepackage[T1]{fontenc}

\usepackage[utf8]{inputenc}

\usepackage{microtype}

\usepackage{inconsolata}

\title{From Ultra-Fine to Fine: Fine-tuning Ultra-Fine Entity Typing Models to Fine-grained}

\author{Hongliang Dai\textsuperscript{1} and Ziqian Zeng\textsuperscript{2} \\
  \textsuperscript{1}College of Computer Science and Technology, \\
  Nanjing University of Aeronautics and Astronautics \\
  \texttt{hongldai@nuaa.edu.cn} \\
  \textsuperscript{2}Shien-Ming Wu School of Intelligent Engineering, \\
  South China University of Technology \\
  \texttt{zqzeng@scut.edu.cn} \\}

\begin{document}
\maketitle
\begin{abstract}
For the task of fine-grained entity typing (FET), due to the use of a large number of entity types, it is usually considered too costly to manually annotate a training dataset that contains an ample number of examples for each type. A common way to address this problem is to use distantly annotated training examples that contains incorrect labels. But the errors in the automatic annotation may limit the performance of trained models.
Recently, there are a few approaches that no longer depend on such weak training data. However, without using sufficient direct entity typing supervision may also cause them to yield inferior performance.
In this paper, we propose a new approach that can avoid the need of creating distantly labeled data. 
We first train an entity typing model that have an extremely broad type coverage by using the ultra-fine entity typing data. 
Then, when there is a need to produce a model for a newly designed fine-grained entity type schema, we can simply fine-tune the previously trained model with a small number of corresponding annotated examples.
Experimental results show that our approach achieves outstanding performance for FET under the few-shot setting. It can also outperform state-of-the-art weak supervision based methods after fine-tuning the model with only a small-size manually annotated training set.
\end{abstract}

\section{Introduction}
\label{sec:intro}

Entity Typing is the task of assigning type labels to entity mentions in texts. 
Its results have been shown to be beneficial to downstream tasks such as Entity Linking \cite{ling2015design,vashishth2021improving}, Coreference Resolution \cite{onoe2020interpretable}, etc. 

Currently, there are mainly two forms of entity typing tasks: Fine-grained Entity Typing (FET) \cite{ling2012fine} and Ultra-fine Entity Typing (UFET) \cite{choi2018ultra,lee2020chinese}. Table \ref{tab:ufet-examples} and Table \ref{tab:fet-examples} provide a few examples for them. The main difference between them lies in the type schemas used. FET uses manually designed type schemas. The entity types are usually organized into a hierarchical structure. UFET directly uses words and phrases as target entity types. This allows it to have a much broader type coverage than an FET task. For example, the UFET dataset constructed in \cite{choi2018ultra} uses a type schema of about 10k types. Moreover, it also uses context dependent types like ``victim'', ``passenger''. 

\begin{table}
\centering
\begin{tabular}{p{4.6cm}p{2cm}}
\hline
\small
\textbf{Sentence with Entity Mention} & \small \textbf{Labels} \\
\hline
\small
Police said \colorbox{yellow!50}{he} had been kidnapped from his home on Tuesday. & \small person, victim, man, male \\
\hline
\small He competed at the 2008 Summer Olympics, where despite missing \colorbox{yellow!50}{the finals} by .13 second, he posted a personal best time. & \small event, match \\
\hline
\small Embassy Suites was owned by \colorbox{yellow!50}{Promus Hotel Corporation}, a hotel management and franchise company from Memphis, Tennessee. & \small company, business, corporation, organization \\
\hline
\end{tabular}
\caption{Examples of Ultra-Fine Entity Typing. Target entity mentions are highlighted with yellow background.}
\label{tab:ufet-examples}
\end{table}

\begin{table}
\centering
\begin{tabular}{p{4.6cm}p{2cm}}
\hline
\small
\textbf{Sentence with Entity Mention} & \small \textbf{Labels} \\
\hline
\small
In the first RTC transaction with a foreign buyer, Royal Trustco Ltd., \colorbox{yellow!50}{Toronto}, will acquire Pacific Savings Bank, Costa Mesa, Calif. & \small /location, /location/city \\
\hline
\small The Fiero plant was viewed as a model of union-management cooperation at GM before slow sales of \colorbox{yellow!50}{the Fiero} forced the company to close the factory last year . & \small /other, /other/product, /other/product/car \\
\hline
\end{tabular}
\caption{Examples of Fine-grained Entity Typing. Target entity mentions are highlighted with yellow background.}
\label{tab:fet-examples}
\end{table}

However, a problem of UFET is that since its entity types are just words or phrases and there are a large number of them, its results are difficult to be exploited in applications. Thus, we believe that in real-world practice, people would still prefer FET in most cases. Therefore, in this paper, FET is our main focus.

For both UFET and FET, it is labor-intensive to manually annotate training examples because of the use of large entity type sets. So far, a commonly adopted approach to address this problem is to use automatically generated weak training data \cite{ling2012fine,choi2018ultra}. The main approach to achieve this is to perform distant labeling with the help of a knowledge base \cite{ling2012fine}. Such generated weak training data are used in most of existing entity typing studies \cite{lin2019attentive,dai2021ultra}. However, the automatically labeled data contains errors. Thus, training the model with them will inevitably limit the final performance. 
Another problem is that, whenever there is a new FET task with a newly designed entity type schema, a new set of training data has to be generated specifically for it. This problem is not trivial since generating training data also requires human effort, and it usually has to be done by an expert. 

Recently, there are a few entity typing studies \cite{ding2021prompt,huang2022few,li2022ultra} that do not rely on creating a weak training dataset for each target entity type schema. For example, both \citet{ding2021prompt} and \citet{huang2022few} propose approaches to learn FET models when there are only a few training examples. \citet{ding2021prompt} employ self-supervision; \citet{huang2022few} use automatic label interpretation and instance generation. However, we think that not using a sufficient amount of entity typing supervision may weaken the capability of the trained models.

Therefore, in this paper, we propose a new entity typing approach that exploits the UFET training data to avoid the requirement of having to create large size weak training data for FET tasks. Since the type schema used by UFET covers a very broad range of entity types, a trained UFET model should contain much helpful information that can benefit different FET tasks, whose type schemas are usually a lot narrower. However, to the best of our knowledge, no existing work has studied to fine-tune a UFET model into an FET model. 

\begin{figure}
    \centering
    \includegraphics[width=50mm]{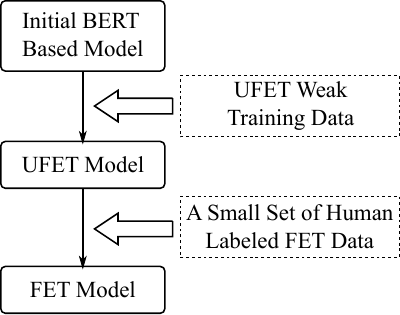}
    \caption{The general procedure of our approach.}
    \label{fig:general-proc}
\end{figure}

The general procedure of our approach is in Figure \ref{fig:general-proc}. First, we train a BERT based entity typing model with UFET training data to obtain a UFET model. This model can be viewed as a pretrained entity typing model and be stored for future use. Whenever there is a new FET task with a newly designed type schema, we can simply fine-tune the trained UFET model with only a small number of corresponding human annotated examples to produce a well-performing model. To better exploit the UFET data for FET, our entity typing model treats type labels as words/phrases that can be tokenized into sequences and then encoded into vector representations. In this way, all the trained parameters of the UFET model can be reused while fine-tuned into an FET model. Moreover, this also allows the model to use the semantic information of the type labels.

We evaluate our approach on commonly used UFET and FET datasets. We first verify that our UFET model achieves favorable performance on the dataset built by \cite{choi2018ultra}. Then, for our main target, FET, on OntoNotes \cite{gillick2014context}, Few-NERD \cite{ding2021few} and BBN \cite{weischedel2005}, our approach yields much better performance than the existing state-of-the-art approach under the few-shot setting. 
Moreover, we also conduct experiments to show that our FET model fine-tuned with only a small set of human labeled data can outperform traditional approaches that use a large set of weak training data.

Our main contributions are summarized as follows.
\begin{itemize}
\item To the best of our knowledge, we are the first to propose fine-tuning UFET models into FET models.
\item We propose an entity typing model that can be better exploited when transferring from UFET to FET.
\item We conduct experiments on both UFET and FET datasets to verify the effectiveness of our approach.
\end{itemize}

Our code is available at \url{https://github.com/hldai/fivefine}.

\section{Methodology}
\label{sec:method}


\subsection{General Procedure}

The general procedure of our approach is illustrated in Figure \ref{fig:general-proc}. Our final target is to obtain models for FET tasks. To this end, first, we train our BERT based entity typing model with Ultra-fine Entity Typing data to obtain a UFET model. 
Note that at this stage, we only use automatically generated weak training examples and do not further fine-tune the model with human annotated UFET data. This is because if the number of manually labeled UFET examples is not large, the generalization ability of the model can be limited after fine-tuning with them.

The obtained UFET model will not be directly used in practice. Instead, it is prepared so that when there is a target FET task, it can be further fine-tuned into a corresponding FET model. In this step, a small number of training examples manually annotated for the target FET task is used to further fine-tune the model.


\subsection{Unifying Predictions for UFET and FET}

\begin{figure*}
    \centering
    \includegraphics{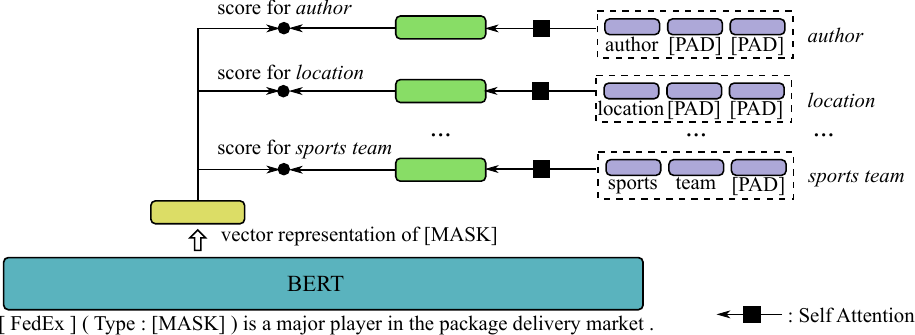}
    \caption{Our entity typing model. Each type word/phrase is tokenized into a sequence. Then, they are padded to same length for convenience of implementation and faster computation.}
    \label{fig:et-model}
\end{figure*}

One main problem in the procedure is how to fine-tune the UFET model into an FET model, since their type schemas are hugely different. A commonly used approach that can achieve this is to simply use a different classification head for the FET model, and only load the parameters of the BERT encoder in the UFET model. However, using a new, untrained classification head loses the type label information learned in the UFET model, and may also make it difficult to exploit the loaded parameters during fine-tuning. 

Using a prompt-based approach \cite{ding2021prompt} is one possible way to better exploit the parameters of a trained UFET model, since the tokens predicted by a Masked Language Model (MLM) can be mapped to the type labels of the target FET task. However, a ``[MASK]'' location only corresponds to one token, which limits the ability of the model to predict multi-word type labels (e.g., \textit{/organization/sports\_team}). Moreover, an MLM is essentially performing multi-class single label classification, while UFET and FET tasks are usually multi-class multi-label classification.



Therefore, we propose a new entity typing model to address the above problems.
The main idea is that we make the model capable of outputting a score when given any entity type word/phrase (Note that this type word/phrase is not necessary from a UFET type schema, or any other type schemas). 
The output score indicates whether this entity type word/phrase is correct for the mention. 
The model itself is ``unaware'' of the existence of type schemas.

Specifically, let $x$ be a target entity mention example, and $t$ be an entity type word/phrase. The model produces a score $s(x,t;\theta)$. With this model, denote $\mathcal{T}_U$ as the type set used by the UFET data in our general procedure, and $\mathcal{T}_F$ as the type set of the target FET task. For UFET, since the types are already words or phrases, the model can directly compute scores for the types in $\mathcal{T}_U$ and thus be trained on the data. 
Benefiting from the broad type coverage of UFET, training the model on the UFET data allows it to learn about a wide variety of both entity mention examples and entity type words/phrases.
For the target FET task, however, the original entity types in $\mathcal{T}_F$ are labels organized into a hierarchical structure instead of words/phrases. To make the model ``recognize'' them more easily, we map each type label $t\in\mathcal{T}_F$ to a type word/phrase $t^*\in\mathcal{T}_F^*$. Then we use $s(x,t^*;\theta)$ as the score for $t$ instead.
For example, the type label \textit{/organization/company} can simply be mapped to the word ``company''. 
Then for FET, the model predicts type words/phrases in ${T}_F^*$ instead of directly predicting labels in $\mathcal{T}_F$. Below are are a few examples of mapping an FET type label to a corresponding word/phrase: \\
\indent \textit{/person/athlete} $\rightarrow$ athlete \\
\indent \textit{/organization/sports\_team} $\rightarrow$ sports team \\
\indent \textit{/other/body\_part} $\rightarrow$ body part \\
It can be seen that the mapping is easy to construct since in most cases we simply use the last part of the type label as its corresponding word/phrase.

\subsection{Entity Typing Model}
\label{sec:et-model}

Our entity typing model is illustrated in Figure \ref{fig:et-model}. For an entity mention in a sentence, we first construct the following sequence and feed it to a BERT encoder: \\
\indent \textit{<lcxt>} \texttt{[}\textit{<mstr>}\texttt{] (Type: [MASK])} \textit{<rcxt>} \\
where \textit{<mstr>} denotes the mention string; \textit{<lcxt>} and \textit{<rcxt>} denote the context text to the left and the right of mention, respectively. For example, the following sentence: \\
\texttt{\colorbox{yellow!50}{FedEx} is a major player in the package delivery market.} \\
where ``FedEx'' is the target mention will be transformed into: \\
\texttt{{[} FedEx {]} (Type: [MASK]) is a major player in the package delivery market.}

Denote the target example (consists of both the target mention and its context) as $x$. 
We feed its corresponding sequence to BERT and obtain the last layer hidden states of the ``[MASK]'' token. Denote this vector as $\bm{h}_x^*\in \mathbb{R}^d$, where $d$ is the hidden size of the BERT model. Then, we apply a transformation to $\bm{h}_x^*$ to get a representation for $x$:
\begin{equation}
    \bm{h}_x=\text{LayerNorm}(f(\bm{h}_x^* \bm{W})),
\end{equation}
where $f$ is a non-linear function; $\bm{W}\in \mathbb{R}^{d\times d}$ is a trainable parameter matrix.

We also obtain a vector representation for each entity type word/phrase. To this end, we first perform tokenization to each type word/phrase. This will result in different lengths of token sequences for different types. During training or evaluation when the target entity type schema is fixed, we pad all these token sequences to same length to avoid having to encode each type separately. Each token is assigned a vector embedding. Specifically, we reuse the weights in the classification head of the BERT masked language model \cite{devlin2019bert} as type token embeddings.

Denote $\bm{X}_t\in \mathbb{R}^{n\times d}$ as the matrix formed with the sequence of embedding vectors corresponding to the sequence of tokens of entity type $t$, where $d$ is the dimension of type token embeddings, $n$ is the sequence length.
We obtain a representation for $t$ by using multi-head self-attention \cite{vaswani2017attention}. Each head has its own sets of trainable parameters $\bm{q},\bm{W}_k,\bm{W}_v$ and computes a vector representation with equation
\begin{equation}
   \text{Attention}(\bm{X}_t)= \text{softmax}(\frac{\bm{q}\bm{K}^T}{\sqrt{d}})\bm{V},
\end{equation}
where $\bm{K}=\bm{X}_t\bm{W}_k,\bm{V}=\bm{X}_t\bm{W}_v$. Then, we use the concatenation of the output vectors of all the heads as the representation for type $t$, denote it as $\bm{g}_t$.

After obtaining $\bm{h}_x$ and $\bm{g}_t$, we use their dot product as the score of type $t$:
\begin{equation}
    s(x,t)=\bm{h}_x \cdot \bm{g}_t
\end{equation}

\subsection{Model Training}

Both UFET and FET tasks are multi-class multi-label classification problems. Thus, we use binary cross-entropy loss to train the model:
\begin{equation}
\label{eq:et-loss}
\begin{split}
    \mathcal{L}_{ET}&=-\frac{1}{|\mathcal{X}|}\sum_{x\in \mathcal{X}}\sum_{t\in \mathcal{T}} [y_{x,t} \cdot \log p(x,t) \\
    &+(1-y_{x,t}) \cdot \log(1-p(x,t))],
\end{split}
\end{equation}
where $\mathcal{X}$ is the training example set; $\mathcal{T}$ is the entity type set used by the entity typing task; $p(x,t)=\sigma(s(x,t))$, $\sigma$ is the sigmoid function; $y_{x,t}$ equals to 1 if $t$ is annotated as a type for $x$ and 0 otherwise.

Although the UFET task covers a huge number of entity types, some of the types may only have a few examples in the training data. As a result, some of the token embeddings of type words/phrases may not get sufficiently trained. Therefore, apart from the entity typing objective, we also use a Masked Language Model objective while training the model with UFET weak training data. We follow the MLM setting in \cite{devlin2019bert} and obtain a corresponding loss based on the token sequence we construct for entity typing in Section \ref{sec:et-model}. Note that the [MASK] token that already exists in the constructed sequence for entity typing is not considered as a masked token slot while computing the MLM loss. With the MLM objective, we make the type token embeddings in our model share the weights as the last linear layer in the MLM classification head. This can help learn better embeddings for type tokens, especially for those that do not occur frequently in the type labels of the training examples.

Another problem the entity typing model faces is that although we surrounded the target entity mention with ``{[}'' and ``{]}'', it can still be difficult for the model to learn to distinguish the mention from the rest of the sentence. Because the supervision signals provided for the model are just entity type labels. Thus, another objective we use for model training is to let the model predict the words immediately to the left and right of the mention. We call this task Neighbor Word Prediction (NWP). To add this objective, for a target example, we first construct a new sequence for feeding to BERT: \\
\indent \textit{<lcxt>} \texttt{[}\textit{<mstr>}\texttt{]} (\textit{<pos>}\texttt{: [MASK])} \textit{<rcxt>} \\
where \textit{<lcxt>}, \textit{<rcxt>} and \textit{<mstr>} are already explained in Section \ref{sec:et-model}; \textit{<pos>} is ``Left'' when predicting the left nearest word (i.e., the last word in \textit{<lcxt>}) and is ``Right'' when predicting the right nearest word (i.e., the first word in \textit{<rcxt>}). To perform prediction, we obtain the last layer hidden states of ``[MASK]'' after feeding the sequence to BERT, and apply a new MLM classification head to it. This MLM classification head used here is different from the one used for the above MLM objective since the two tasks are different. We also use cross entropy loss for NWP.

Let $\mathcal{L}_{MLM}$ be the loss for the MLM objective, and $\mathcal{L}_{NWP}$ be the loss for the NWP objective. Then, while training our entity typing model with the weak UFET data, we use the following final loss to perform multi-task learning:
\begin{equation}
\label{eq:full-loss}
\mathcal{L}=\mathcal{L}_{ET}+\lambda_{MLM}\mathcal{L}_{MLM}+\lambda_{NWP}\mathcal{L}_{NWP},
\end{equation}
where $\lambda_{MLM}$ and $\lambda_{NWP}$ are two hyperparameters controlling the strengths of the MLM and the NWP objectives, respectively.

For the UFET task, we follow \cite{dai2021ultra} and train our model with the full training data they created. Smaller weights are also assigned for labels generated through prompting in the loss since they are less accurate.

When fine-tuning the trained UFET model for FET tasks, we directly use the loss $\mathcal{L}_{ET}$ in Equation \ref{eq:et-loss}, since there are not so much training data.

\section{Related Work}

For both UFET and FET, due to the use of large entity type sets, it is labor-intensive to manually annotate training examples. Thus, different approaches \cite{ling2012fine,choi2018ultra,dai2021ultra} of automatically generating weakly labeled training examples are proposed. Among them, the most commonly used method is to link entity mentions to a knowledge base, and then use the types of the corresponding entities as labels \cite{ling2012fine,gillick2014context,choi2018ultra}. Additionally, \citet{choi2018ultra} propose to use the head word of the mention phrase as its type label. \citet{dai2021ultra} generate entity type labels for mentions with a prompt-based method.

With different ways to create large amounts of training data automatically, the incorrectness of the generated labels become a problem. Many entity typing studies \cite{ren2016label,chen2019improving,pang2022divide} seek to obtain better models when using weak training data. For example, \citet{onoe2019learning} learn a neural model to correct noisy entity type labels and filter unuseful examples. \citet{pang2022divide} learn a backbone model as a feature extractor and a noise estimator, and perform feature cluster based loss correction afterwards.

Recently, there are more entity typing studies that do not follow the commonly adopted approach of training with distantly labeled data created by using a knowledge base. 
Some of them also do not require a designated training set for each entity type schema. For example,
\citet{li2022ultra} exploit indirect supervision from natural language inference. \citet{ding2021prompt} employ self-supervision instead of explicit type labels. 
\citet{huang2022few} use automatic label interpretation and instance generation to achieve few-shot FET.

\section{Experiments}

We conduct experiments on both UFET and FET datasets. In this section, we use \textbf{FiveFine} to denote our approach (Because there are five ``fines'' in the title of this paper).

\subsection{Datasets}

For UFET, we use the dataset built by \citet{choi2018ultra}, which to the best of our knowledge, is the only English UFET dataset that is publicly available. Its target entity type set contains 10,331 types that are all free-form words or phrases. Apart from a broad type coverage, it also uses various forms of entity mentions, including named entity mentions like ``Joe Biden'', pronoun mentions like ``she'', and nominal mentions like ``the nearby university''. Thus, it is very suitable to be used to train an entity typing model that can be further fine-tuned for specific FET tasks. This dataset contains more than 20M distantly labeled training examples and 6,000 manually annotated examples evenly split into train, dev and test. In addition, we also use the labels generated by \citet{dai2021ultra} through prompting, as well as the 3.7M pronoun mention examples they produce.

For FET, we use OntoNotes \cite{gillick2014context}, Few-NERD \cite{ding2021few} and BBN \cite{weischedel2005}.

\begin{itemize}
    \item{\textbf{OntoNotes}} The OntoNotes dataset uses an ontology that consists of 89 entity types. We follow \cite{huang2022few} and use the version that contains 8,963 test examples and 2,202 dev examples. Both the test examples and the dev examples are manually annotated. For training data, we use a version provided by \cite{choi2018ultra}, which contains about 0.8M instances. OntoNotes treats entity typing as a multi-label classification problem. This means that an entity mention can be assigned labels of different type paths. For example, a university can be assigned both \textit{/organization}, \textit{/organization/university} and \textit{/location}. 
    \item{\textbf{Few-NERD}} The Few-NERD dataset uses 66 entity types. We use the supervised setting whose train, dev and test sets contain about 131K, 18K and 37K examples, respectively. All these examples are manually annotated. Unlike OntoNotes, Few-NERD treats entity typing as a single-label classification problem, which means only one fine-grained type can be assigned to a mention. For example, a university can be either assigned \textit{/organization/university} or \textit{/location}. 
    \item{\textbf{BBN}} The BBN dataset uses 46 entity types. We use the version provided by \citet{huang2022few}, whose train, dev and test sets contain about 84k, 2k, 13k examples, respectively. 
\end{itemize}

These datasets will be further processed when used for conducting few-shot FET experiments.

\subsection{Experimental Settings}

For BERT, we use both bert-base-cased and bert-large-cased provided by Hugging Face\footnote{https://huggingface.co/} to train separate entity typing models. 

When training the UFET model, since we mainly follow the training procedure of \cite{dai2021ultra} most of the hyperparameters are set to be same as them. Except for $\lambda_{MLM}$ and $\lambda_{NWP}$, which are new in our approach. We set both of them to 0.1.
Adam is used as the optimizer for all the training. 

In terms of evaluation metrics, we follow existing work. While evaluating the UFET model, we use macro-averaged precision, recall, and F1 \cite{choi2018ultra}. While evaluating the FET models, we use strict accuracy, micro-averaged F1 and macro-averaged F1.

\subsection{UFET Evaluation}
\label{sec:ufet-eval}

Although FET is our main target, we still need to verify that our UFET model performs well. Since otherwise, it may leads to inferior results after fine-tuned to FET.

For UFET, we compare with the following existing methods:
\begin{itemize}
    \item \textbf{MLMET} \cite{dai2021ultra} introduces extra entity typing labels that are generated through prompting. It first trains the entity typing model with weakly labeled data, then conduct self-training with both human annotated data and weak training data. The training procedure of our UFET model also follows MLMET.
    \item \textbf{LITE} \cite{li2022ultra} uses indirect supervision from natural language inference (NLI) to train entity typing models. A problem with this approach is that for each entity mention, the model has to evaluate an NLI example for every entity type. This leads to a very long inference time.
    \item \textbf{MCCE} \cite{jiang2022recall} adopts the cross-encoder based architecture which concatenates the mention with each type and feeds the pairs into a pretrained language model. It speeds up inference with a recall-expand-filter paradigm. This approach currently yields the best performance on the UFET dataset created by \cite{choi2018ultra}.
    \item \textbf{Box} \cite{onoe2021modeling} captures latent type hierarchies with box embedding. 
    \item \textbf{BERT-Direct} directly trains a BERT-Based model by using the human annotated data. The model feeds \texttt{[CLS]} \textit{<sentence>} \texttt{[SEP]} \textit{<mstr>} \texttt{[SEP]} to BERT and use the output vector of the [CLS] token for classification.
\end{itemize}
\begin{table}
\centering
\begin{tabular}{lccc}
\hline \textbf{Method} & \textbf{P} & \textbf{R} & \textbf{F1} \\ \hline
BERT-Direct & 51.0 & 33.8 & 40.7 \\ 
MLMET & 53.6 & 45.3 & 49.1 \\
LITE & 52.4 & \textbf{48.9} & 50.6 \\ 
MCCE & \textbf{56.3} & 48.5 & \textbf{52.1} \\ 
Box & 52.8 & 38.8 & 44.8 \\ 
\hline
FiveFine-Base (No MLM) & 49.3 & 48.5 & 48.9 \\
FiveFine-Base (No NWP) & 53.7 & 46.3 & 49.8 \\
FiveFine-Base & 53.7 & 47.3 & 50.3 \\
FiveFine-Large & 53.0 & 48.6 & 50.7 \\ \hline
\end{tabular}
\caption{\label{tab:ufet-perf} Macro-averaged Precision, Recall, and F1 of different approaches on the UFET dataset. FiveFine-Base and FiveFine-Large are our models based on BERT-Base and BERT-Large, respectively. FiveFine-Base (No MLM) and FiveFine-Base (No NWP) and our models trained without the MLM objective and without the NWP objective, respectively. }
\end{table}

\begin{table*}
\centering
\begin{tabular}{lccccccccc}
\hline & \multicolumn{3}{c}{\textbf{OntoNotes}} & \multicolumn{3}{c}{\textbf{Few-NERD}} & \multicolumn{3}{c}{\textbf{BBN}} \\
\hline \textbf{Method} & \textbf{Acc} & \textbf{MiF1} & \textbf{MaF1} & \textbf{Acc} & \textbf{MiF1} & \textbf{MaF1} & \textbf{Acc} & \textbf{MiF1} & \textbf{MaF1} \\ \hline
BERT-Direct & 17.15 & 37.38 & 41.50 & 29.43 & 39.22 & 39.22 & 5.11 & 25.0 & 24.7 \\
ALIGNIE & 60.74 & 75.08 & 76.38 & 57.45 & 69.54 & 69.54 & 71.33 & 77.78 & 76.50 \\ \hline
FiveFine & \textbf{65.59} & \textbf{83.66} & \textbf{85.42} & \textbf{61.22} & \textbf{71.88} & \textbf{71.88} & \textbf{75.00} & \textbf{81.08} & \textbf{80.71} \\ \hline
\end{tabular}
\caption{\label{tab:fet-perf} FET performance under the 5-shot setting. Due to the use of different randomly sampled train and dev examples, results for ALIGNIE are different from those reported in \cite{huang2022few}. ``MiF1'' means micro-averaged F1; ``MaF1'' means macro-averaged F1.}
\end{table*}

For our approach, we report the results of both models based on BERT-Base and BERT-Large, which are represented with \textbf{FiveFine-Base} and \textbf{FiveFine-Large}, respectively. 
In addition, for FiveFine-Base, we also report the performances when trained without the MLM objective and without the NWP objective. They are represented with \textbf{FiveFine-Base (No MLM)} and \textbf{FiveFine-Base (No NWP)}, respectively.

The results are in Table \ref{tab:ufet-perf}. Our model based on BERT-Large only fails to beat the most recent approach MCCE.
The favorable performance of our model indicates that it has exploited the UFET training data well, which we believe would help it to achieve good performance after being fine-tuned for specific FET tasks.


Comparing FiveFine-Base, FiveFine-Base (No MLM) and FiveFine-Base (No NWP), first, we can see that the performance of our model drops when trained without the MLM objective. This verifies the benefit of including it in the training loss. We think MLM helps to learn better type token embeddings, since they share the same weights as the final linear layer of the MLM classification head. But the decrease in performance is much less significant when the NWP objective is removed. We think the reason is that since NWP only requires to predict the neighboring words, the help it provides for the model to learn that the entity mentions are the targets to be classified is limited. 


\subsection{FET Evaluation}

For evaluation on FET, we mainly follow the setting in \cite{huang2022few} to evaluate our approach under the few-shot setting.

For OntoNotes and BBN, same as \cite{huang2022few}, we filter the entity types that do not contain enough instances to form few-shot datasets. Afterwards, 21 types for OntoNotes and 25 types for BBN remain. We also follow the code released by \citet{huang2022few} to process the test sets, which further filters some examples that their approach has difficulty dealing with (e.g., examples labeled with multiple type paths). This results in 3,461, 95,880 and 12,258 test instances remaining for OntoNotes, Few-NERD and BBN, respectively.


For each dataset, we sample examples to build 5-shot train and dev sets. Both the train and the dev sets contain 5 examples for each entity type. We repeat five experiments for each dataset and report the average results. Each time, different train and dev sets are randomly sampled.

The following methods are compared:
\begin{itemize}
    \item \textbf{ALIGNIE} \cite{huang2022few} is the state-of-the-art approach for FET under the few-shot setting. It uses a type label interpretation module to learn to relate types labels to tokens, and an instance generator to produce new training examples.
    \item \textbf{BERT-Direct}: Same as the BERT-Direct model in Section \ref{sec:ufet-eval}.
\end{itemize}


Note that the results for ALIGNIE will be different from those reported in \cite{huang2022few}. Because the 5-shot data are randomly sampled by us, and the OntoNotes training data we use are also different from theirs.

For our approach, we fine-tune the FiveFine-Base model with the few-shot FET training data. 

Table \ref{tab:fet-perf} presents the results. FiveFine achieves the best performance on all three datasets. Especially on OntoNotes and BBN, it outperforms ALIGNIE by a large margin. We think this is because the quality of the weak training data of OntoNotes and BBN is not good. As a result, ALIGNIE is not able to learn a well performing model from them. But since our model is pretrained with UFET data, the model itself already possesses the power to do entity typing before it is fine-tuned on the few-shot data. This allows it to produce much better results when the training data are of bad quality. In addition, we believe the quality of the training data is also a main reason why BERT-Direct performs poorly.

\subsection{Comparing Weak Supervision and Human Annotation}

\begin{table}
\centering
\begin{tabular}{lccc}
\hline \textbf{Method} & \textbf{Acc} & \textbf{Micro-F1} & \textbf{Macro-F1} \\ \hline
MLMET & 67.4 & 80.4 & 85.4 \\
ANL & 67.8 & 81.5 & 87.1 \\
\hline
BERT-Direct & 50.1 & 67.8 & 74.6 \\
FiveFine & \textbf{69.3} & \textbf{84.8} & \textbf{89.4} \\ \hline
\end{tabular}
\caption{\label{tab:fet-human-perf} Performance of FiveFine and BERT-Direct on OntoNotes trained with a small-size human annotated dataset comparing with approaches that use large size weak training data.}
\end{table}

We also compare the performance of our FET model that is fine-tuned with only a small set of human labeled data against traditional approaches that use a large set of weak training data. To this end, we perform human annotation for the OntoNotes dataset by using the examples from its training and dev set. For each type, we first select at most 100 candidate examples, and then ask the annotator to go through the examples and find at most 10 correct ones. While selecting the 100 candidate examples, we try to keep the word overlap number of different examples small to ensure variety. We also randomly select at most 5 examples for each type from the original dev set to produce a small sized new dev set. In this way, we collect 675 training examples. Note that this constructed data do not strictly follow the few-shot setting, because some of the types would have less than 10 training examples. 

We compare with weak supervision based approaches MLMET \cite{dai2021ultra} and ANL \cite{pan2022automatic}. ANL is a state-of-the-art approach that trains the model after automatically correcting the noisy labels. Both MLMET and ANL are trained with the original full distantly labeled data.

Apart from our approach, we also train BERT-Direct with the manually annotated data we create and report its performance.

The results are in Table \ref{tab:fet-human-perf}. By using only a small number of training examples, FiveFine already outperforms the compared methods. This verifies that instead of creating large size weak training data, it can be more preferable to use our approach to produce FET models with small human labeled datasets.

\section{Conclusion}

In this paper, we propose the approach to fine-tune a UFET model to FET models, which can avoid the requirement of constructing distantly labeled training data when an application needs to train a model for a newly designed FET type schema. 
This approach is feasible because the type schema used by UFET have very broad type coverage, usually much broader than FET tasks. 
We also propose an entity typing model that treats target entity type labels as words/phrases. This allows all the trained parameters of the model to be reused when fine-tuned from UFET to FET, so that the trained UFET model can be better exploited. The experiments we conduct verify the effectiveness of both our UFET model, and the FET models that are fine-tuned from it with small sized training sets.

\section*{Limitations}
We train a UFET model and then fine-tune it for target FET tasks. In our approach, the UFET training data is the main source of limitations. First, the large size UFET training data are automatically generated, and thus may contain errors. Such errors can propagate to the fine-tuned FET models. Another problem is that, for some entity types, there are not many training examples.
Moreover, some types useful in specific domains (e.g., adverse drug reaction for the biomedical domain) are not included in the UFET type vocabulary at all. As a result, the UFET model will not be as helpful when applied to FET data that contain such types.

\section*{Acknowledgements}
The authors would like to thank the reviewers for their insightful comments and suggestions.

\bibliography{custom}
\bibliographystyle{acl_natbib}




\end{document}